\pdfoutput=1
\documentclass[11pt]{article}

\usepackage{acl}

\usepackage{times}
\usepackage{latexsym}

\usepackage[T1]{fontenc}
\usepackage[utf8]{inputenc}

\usepackage{microtype}

\usepackage{inconsolata}

\usepackage{lipsum}

\usepackage{graphicx}

\newenvironment{LLM}{\ttfamily\small\setlength{\parindent}{0cm}\vspace*{0.5\baselineskip}\setlength{\parskip}{0.2\baselineskip}}{\vspace*{0.2\baselineskip}}

\newenvironment{LLMTrans}{\small\itshape\setlength{\parindent}{0cm}(}{)}

\title{Child vs. machine language learning:\\Can the logical structure of human language unleash LLMs?}

\author{Uli Sauerland \\
  ZAS \\
  \texttt{\small sauerland@leibniz-zas.de} \\\And
  Celia Matthaei \\
  ZAS \& HU Berlin \\
  \texttt{\small celia.matthaei@gmail.com} \\\And
  Felix Salfner\\
  independent researcher\\
  \texttt{\small research@felix.salfner.de}}

\begin{document}
\maketitle
\begin{abstract}

We argue that human language learning proceeds in a manner that is different in nature from current approaches to training LLMs, predicting a difference in learning biases. We then present evidence from German plural formation by LLMs that confirm our hypothesis that even very powerful implementations produce results that miss aspects of the logic inherent to language that humans have no problem with. We conclude that attention to the different structures of human language and artificial neural networks is likely to be an avenue to improve LLM performance.

%Artificial neural networks are currently implemented with a fixed number of unit layers, e.g. GPT 4.0 has 125 layers. We argue that human language learning proceeds in a manner that would not be predicted by such a fixed layer structure. We then present evidence from German plural formation by LLMs that the prediction is born out: even very powerful implementations produce results that are distinct from humans in the way the rigid layer structure predicts. We conclude that a more flexible network organization in artificial neural networks is likely to be an avenue to improve LLM performance.
\end{abstract}

\section{Introduction}

Results from how humans learn language suggest that logical connections drive learning strongly.
In particular, \citet{sauerlandetal24} point out that children learning language are spontaneously creating the negation of an output.
Their example is based on the German prepositions \textit{mit} (`with') and \textit{ohne} (`without'). Like their English translations the two are logical contraries, but unlike in English this is not morphologically transparent via the morphological complexity of with-out. The surprising observation is that children learning German spontaneously and frequently produce sequences \textit{mit ohne} (lit. `with with-out'), that are not fully grammatical and exceedingly rare in the adult language.
This finding argues that after learning to output \textbf{A} in some circumstances, the negative output \textbf{\=A} `not A' is readily available to the human learner and German children therefore are drive to produce the inverse of \textit{mit} as two words, \textit{mit} and a marker of negation.
Other findings from child language acquisition \cite{cesana-arlotti18,guastietal23} support this conclusion, and recent evidence shows that other primates also use negation in concept representation \cite{dautricheetal22}.

%Neural networks can represent logical/algebraic operations such as negation, but the logical way to do so requires an additional layer in a network. Assume that one unit a of a neural network yields output A. There are two ways to add to the network what amounts to the negation of A (see figure 1). 1) Add a unit \textbf{\=a} that receives the same inputs as X, but with inverse weights or activation function, to produce output \textbf{\=A}. 2) Move unit \textbf{a} to the penultimate layer and add units \textbf{=} and \textbf{$\neg$} that receive the output of \textbf{a} as  input with positive and negative weights respectively to yield outputs \textbf{A} and \textbf{\=A}. Since way 1) would require independent learning of \textbf{\=a} and not inherently capture the logical relationship between A and \=A, we put it aside for the following. The second way captures the logical relationship, but crucially involves an extra layer in a rigidly layered network. But if a network has learned to connect unit \textbf{a} with output \textbf{A} by backward induction, \textbf{a} would be a unit in the output layer of the network. This predicts that learning logical negation of \textbf{A} will not be straightforward for the network because connecting units of the output layer to other units of the network breaks the feed-forward topology. As a result, we expect LLMs to disprefer learning logical connections over other generalizations.

\begin{figure}
    \centering
    \begin{tabular}{lll}\llap{a.}
           \includegraphics[width=0.25\linewidth]{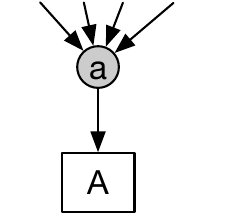}  &  b.
           \includegraphics[width=0.25\linewidth]{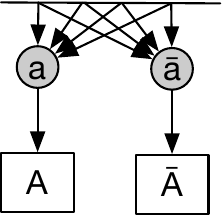} & c.
           \includegraphics[width=0.25\linewidth]{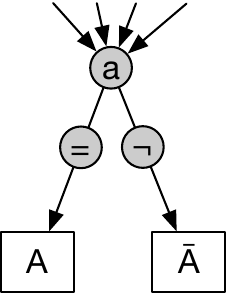}\\ 
    \end{tabular}
    \caption{Implementation of two ways of adding the negation of A in a neural
    network. In panel a, neuron \textbf{a} yields
    output A. Panel b and c illustrate two possible ways of obtaining output \=A whenever A is not outputted. In panel b, an independent neuron
    \textbf{\=a} with inverse weights output \=A.  In c, two new neurons \textbf{=} and
    $\mathbf{\neg}$ trigger A  and \=A  depending on the output
    of unit \textbf{a}.}
    \label{fg:layers}
\end{figure}

In contrast to the conceptual representation present in humans, neural networks rely heavily on independently inferring concepts from training data. LLM architecture does not have a bias for specific algebraic or logical relationships such as negation of \textbf{A} and \textbf{\=A}. This predicts that learning logical negation of \textbf{A} will not be straightforward and we expect LLMs to not prefer learning logical connections over other generalizations. 
As an example, to model the logical operation of negation of output \textbf{A} produced by neuron \textbf{a}, a second neuron \textbf{\=a} could be added where the two neurons' weights are inversely linked, i.e., shared ($\mathbf{w_{\overline{a}}} = -\mathbf{w_a}$), as is depicted in Figure~\ref{fg:layers}-b. Another approach would be to add a layer to the network by moving unit \textbf{a} to the penultimate layer and add units \textbf{=} and \textbf{$\neg$} that receive the output of \textbf{a} where \textbf{$\neg$} has an inverse activation function to yield \textbf{A} and \textbf{\=A} (see Figure~\ref{fg:layers}-c). But both of these ways involve substantial changes to the network architecture, and we therefore predict that an LLM should require a lot of learning data to learn the complementary distribution of \textbf{A} and \textbf{\=A}.  We conducted experiments that corroborate this prediction, and which are presented below.

Current approaches rely on reinforcement learning with a set of mathematical and logical reasoning exercises, e.g. \cite{deepseek-aietal25}. This training leads to improved performance with tasks involving explicit logical reasoning. But the logical reasoning we refer to is implicit and detected by tasks such as the two-year-olds spontaneously recognizing the logical relationship between \emph{with} and \emph{without} mentioned above. Modeling logical relationships explicitly may help with this, but we suspect may not be sufficient.
The collaboration of computer scientists, linguists and cognitive scientists will likely lead to ideas for the further improvement of LLMs.

\section{A Prediction}

The human drive towards capturing a logical relationship is not predicted to be present in LLMs with a strict layer structure as we argued above. Given enough input LLMs can learn the antonymic contrary relation between \textit{mit} and \textit{ohne} just as well as that between with and with-out. But we argue now that for different language properties, the relative difficulty of learning logical relationships yields sub-human performance.  We focus on the formation of German nominal plurals. German plural formation is based on several noun classes determined by phonological and semantic criteria, but also a default. Using a default requires logical negation – if a noun fits none of the other classes, it belongs to the default class.  We adopted a task from \citep{marcusetal95} that involves the formation of plural forms for nonce nouns, i.e. hypothetical noun stems that don’t actually exist in the language.  

\begin{figure}
    \centering
    \includegraphics[width=0.8\linewidth]{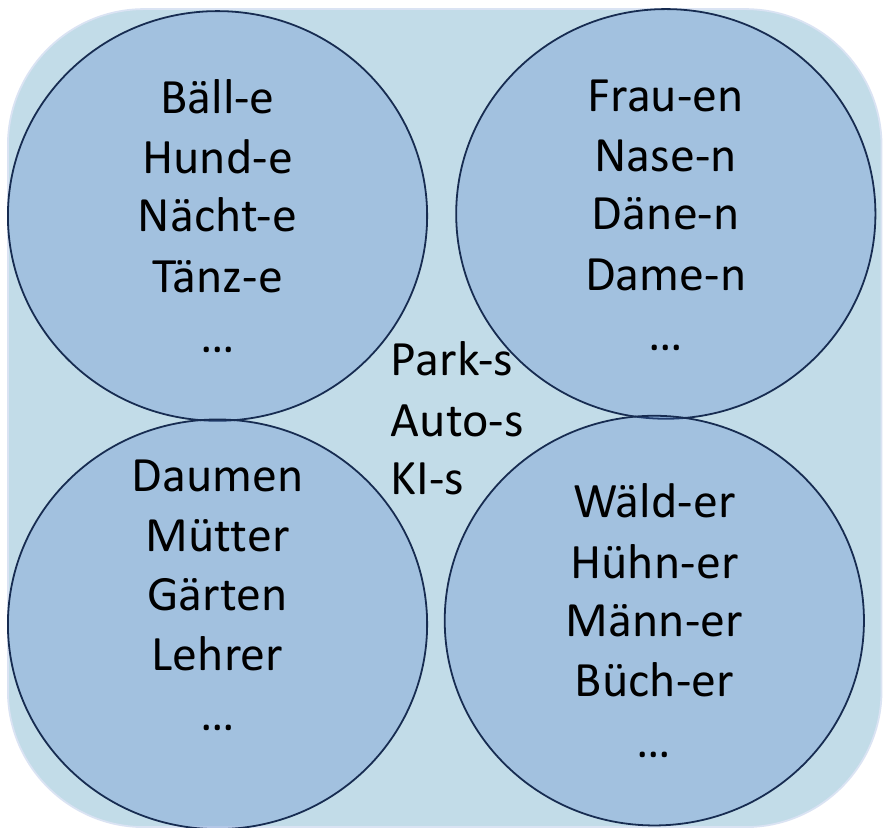}
    \caption{Schematic view of the German plural endings: The majority of noun stems select one of the four irregular plural endings \emph{-e}, \emph{-n}, \emph{-r}, and \emph{-$\emptyset$}, while the default ending \emph{-s} is rare.}
    \label{fg:dplural}
\end{figure}

The formation of plural forms in German has a logical structure that involves a default form, namely the plural ending \emph{-s}.
To illustrate the concept of \emph{default} (or \emph{elsewhere}, \citealt{kiparsky73}) in morphology consider English plurality, where \emph{-s} is also the default form.
A few nominal stems of English such as \emph{child} and \emph{ox} allow a different plural -- \emph{child-ren} and \emph{ox-en} respectively -- and with these forms the use of the default \emph{-s} is ungrammatical: \emph{*child-s} and \emph{*ox-s}. The main difference between German and English is that irregular plurals only occur with a small finite number of stems in English, while in German the default plural only occurs rarely (see Figure~\ref{fg:dplural}).

Morphological systems in language almost always involve defaults as was first observed by the Sanskrit grammarian P\=anini (\citealt{bobaljiksauerland18} and others). 
The logic behind a default involves negation: If a more specific form is available (e.g. \emph{child-ren}) that default form cannot be applied (e.g. *\emph{child-s}).
Our heuristic in the previous section led us to conclude that human language learners should more readily learn generalizations involving negation than current LLMs.
In English, an LLM is still expected to exhibit human-like performance because the overwhelming frequency of \emph{-s} as a plural ending would allow an LLM to recognize the generalization.
In German, on the other hand, the default \emph{-s} plural occurs rarely as noted already above, both by lemma and token occurrence \citep{marcusetal95}.
Therefore we expect an LLM to recognize some subgeneralizations in the more frequent classes. The default nature of the \emph{-s} ending, however, depends on the recognition that the forms with the \emph{-s} plural derive by negating  the four other classes.
This type of generalization would therefore be one humans are prone to notice, while for LLMs they are more difficult.

\section{Testing the Prediction}

To test the prediction, we repurposed the materials of \cite{marcusetal95} into a test for current LLMs.
They created 24 nonce nouns; i.e. nouns that are not actually words of German such as \emph{Bral}, \emph{Klot} and \emph{Fneik} and collected data on how natural different plural forms such as \emph{Bral-e}, \emph{Bral-en}, \emph{Bral-er}, and \emph{Bral-s} feel to German native speakers.
This task reflects the natural, frequent process of new nouns entering the language as a loan from a foreign language, as a product name or via another word-creation process and reflects part of the speakers knowledge of German.
We set out to probe seven current LLM-based chatbots that we had access to on German plural formation.
To test whether LLMs possess human-like knowledge,
we  queried each LLM for each of the 24 nonce nouns of \cite{marcusetal95}  with the German question \emph{Was w\"are der Plural von XXX?} (`What would be the plural form of XXX?') where XXX would be replaced with the nonce noun. For each query, we started a new session with the chatbot. In addition, we posted at most one follow up question within the same session to the chatbot.

We extracted from the dialogues the plural forms the LLMs created from these answers. Then we assigned them a human plausibility rating in the following way:
If the forms conformed to one of the five regular pluralization processes of German morphology, we assigned them the plausibility rating that \cite{marcusetal95} reported from German speakers' judgments for this form.
For entirely impossible plural forms, which \cite{marcusetal95} did not test in their study, we assigned a plausibility of zero.

\begin{figure*}
    \centering
    \includegraphics[width=0.9\linewidth]{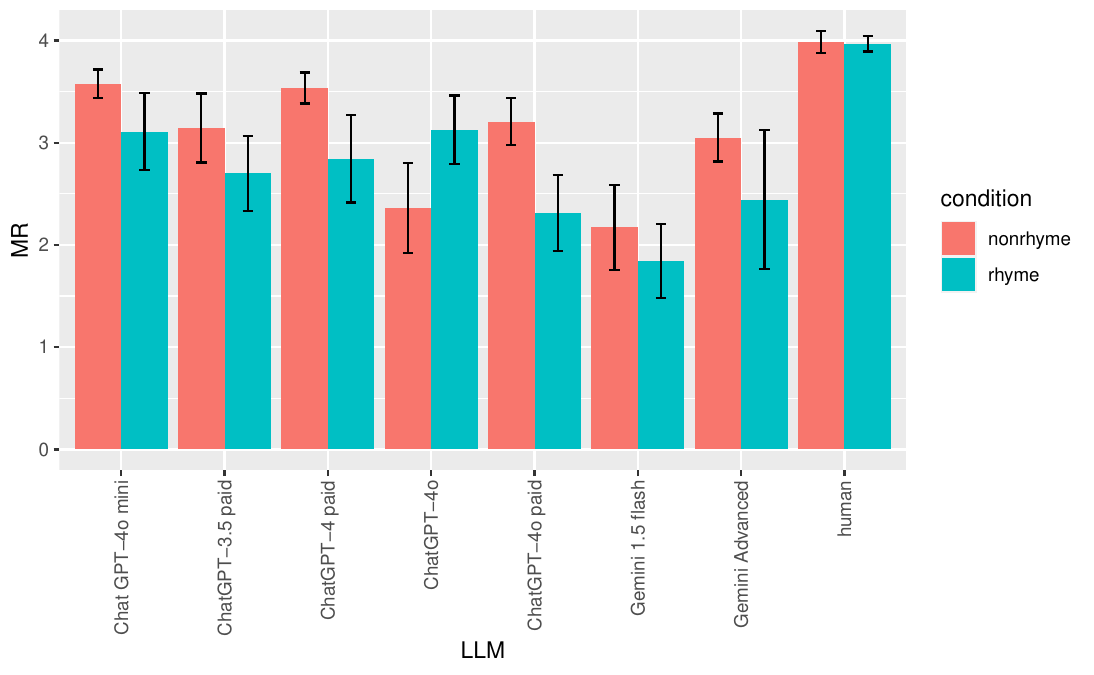}
    \caption{Mean plausibility ratings (MR) of the plural forms produced by LLMs for the German nonce noun task (Marcus et al. 1995) grouped by nouns rhyming with a real German noun and non-rhyming nonce nouns. The rightmost two bars show the human plausibility ratings of the correct plural forms  }
    \label{fig:llmcomparison}
\end{figure*}

The method of data collection sometimes involved a follow-up question.
Namely, if the chatbot seemed to treat our initial as involving a word from another language or the response related to an orthographically similar real German word, our follow-up was \emph{Ich meine das deutsche Wort XXX} (`I mean the German word XXX').
The following transcript shows one sample session of this type:\footnote{Because the Gemini 1.5 Flash chatbot on this trial responded in English, we don't translate the German parts of the interaction into English.  The translations of German examples includes were part of the Chatbot output.  All other times, the chatbots responded in German to the German question.}

\begin{LLM}
\textbf{EXP:} Was ist der Plural von Pisch?

\textbf{Gemini 1.5 Flash:} The noun "Pisch" is an uncountable noun in Plautdietsch, which means that it does not have a plural form. It is always used in the singular form to refer to urine.

Uncountable nouns are typically used to refer to substances, concepts, or things that are considered to be a whole or a mass rather than individual units. For example, other common uncountable nouns in English include "water," "air," "love," and "money."

In some cases, uncountable nouns can be used in a countable sense when referring to specific instances or amounts. For example, you could say "I drank three glasses of water" or "I spent a lot of money on clothes." However, in these cases,the noun is still technically uncountable and does not have a true plural form.

So, the answer to your question is that there is no plural form of "Pisch" in Plautdietsch. The noun is always used in the singular form to refer to urine

\textbf{EXP:} Ich meine das deutsche Wort Pisch

\textbf{Gemini:} I apologize for the error in my previous response. I am still under development, and I am always learning new things. I will try to be more careful in the future.

The German word "Pisch" is a colloquial term for urine, and it is also considered an uncountable noun. This means that it does not have a plural form and is always used in the singular form to refer to urine.

Here are some examples of how to use "Pisch" in a sentence:

"Das Kind hat in die Hose gepischt." (The child peed in his pants.)

"Ich muss dringend mal pischen." (I really need to pee.)

"Der Geruch von Pisch lag in der Luft." (The smell of urine was in the air.)

As with other uncountable nouns, "Pisch" can sometimes be used in a countable sense to refer to specific instances or amounts. For example, you could say "Ich habe drei Pischflecken auf dem Teppich gesehen" (I saw three urine stains on the carpet). However, in these cases, the noun is still technically uncountable and does not have a true plural form.

I hope this helps! Let me know if you have any other questions.
\end{LLM}

In the following case, it was unclear whether the chatbot treated \emph{Pund} to be a misspelling of \emph{Pfund} or as a real word of German.  But since \emph{Pfund} is actually not a \emph{plurale tantum} noun and can combined with the numeral \emph{one}, the chatbot seems to treat \emph{Pund} as a real word of German.

\begin{LLM}
\textbf{EXP:} What is the plural of Pund?

\textbf{Gemini 1.5 flash:} Das Wort "Pund" ist ein Pluraletantum, d.h. es hat keine Einzahlform. Es wird immer in der Pluralform verwendet, unabhängig davon, wie viele Einheiten gemeint sind.

\end{LLM}

\begin{LLMTrans}
The word "Pund" is a Pluraletantum, i.e. it has no singular form. It is always used in the plural form, regardless of how many units are meant. 
\end{LLMTrans}

\begin{LLM}
Die Bedeutung von "Pund" kann je nach Kontext variieren. Es kann sich um eine Gewichtseinheit (ca. 500 Gramm) oder eine Währungseinheit (z.B. britisches Pfund) handeln.

\end{LLM}

\begin{LLMTrans}
The meaning of "Pund" may vary depending on the context. It can be a unit of weight (approx. 500 grams) or currency (e.g. British pound). 
\end{LLMTrans}

\begin{LLM}
Beispiele:

Ich habe mehrere Pund Fleisch gekauft. (Gewichtseinheit)

Der Preis des Computers beträgt 500 Pund. (Währungseinheit)

Wie viele Pund wiegst du? (Gewichtseinheit)

Ich habe ihm zehn Pund geliehen. (Währungseinheit)

Es ist wichtig zu beachten, dass "Pund" keine Einzahlform hat. Die Verwendung von "ein Pund" ist grammatikalisch falsch.

\end{LLM}

\begin{LLMTrans}
It is important to note that "Pund" is not a singular form. The use of "a pound" is grammatically incorrect. 
\end{LLMTrans}

\begin{LLM}
Weitere Informationen zum Wort "Pund" finden Sie in Online-Wörterbüchern wie Duden.de oder Wiktionary.de.

\end{LLM}

\begin{LLMTrans}
Find more information about the word "Pund" in online dictionaries such as Duden.de or Wiktionary.de.
\end{LLMTrans}

The other follow-up question we used was \emph{Bist Du sicher?} (`Are you sure?').  We used this when the initial answer was short and seemed inconsistent, e.g. the use of both \emph{Raune} and \emph{Raunen} in the following.

\begin{LLM}
\textbf{EXP:} Was ist der Plural von Raun?

\textbf{Chat GPT 4o paid:} Der Plural von "Raun" im Deutschen ist "Raune". Es wird meist in der Form "Raunen" verwendet, wenn von leisen, geheimnisvollen Geräuschen oder Stimmen die Rede ist.

\end{LLM}

\begin{LLMTrans}
The plural of "Raun" in German is "Raune". It is usually used in the form of "Raunen" when talking about quiet, mysterious noises or voices.    
\end{LLMTrans}

\begin{LLM}
\textbf{EXP:} Bist du sicher?

\textbf{Chat GPT:} Entschuldige, das war ein Fehler in meiner vorherigen Antwort. Das Wort "Raun" hat tatsächlich keinen Plural, da es nicht existiert. Wenn du "Raunen" meinst, dann ist es bereits die Pluralform des Substantivs "Raun", das allerdings kaum genutzt wird. "Raunen" bezieht sich auf leises Murmeln oder Flüstern.

\end{LLM}

\begin{LLMTrans}
Sorry, that was a mistake in my previous answer. The word "Raun" actually has no plural because it doesn't exist. If you mean "Raunen", then it is already the plural form of the noun "Raun", which, however, is hardly used. "Raunen" refers to quiet murmuring or whispering.    
\end{LLMTrans}

We used the \emph{`Are you sure?'} most frequently with the 4o versions of ChatGPT and in the following case it actually led to a correction from an impossible form to best possible one \emph{-e} ending with a 4.1 rating.

\begin{LLM}
\textbf{EXP:} Was ist der Plural von dem Wort Klot?

\textbf{Chat GPT 4o:} Der Plural des Wortes "Klot" im Deutschen ist "Klötze".

\end{LLM}

\begin{LLMTrans}
The plural of the word "Klot" in German is "Klötze".    
\end{LLMTrans}

\begin{LLM}
\textbf{EXP:} Bist du sicher?

\textbf{Chat GPT:} Entschuldigung, ich habe einen Fehler gemacht. Der richtige Plural von „Klot“ ist tatsächlich „Klote“. Vielen Dank, dass du nachgefragt hast!

\end{LLM}

\begin{LLMTrans}
Sorry, I made a mistake. The correct plural of "Klot" is actually "Klote". Thank you for asking!    
\end{LLMTrans}

But with a different version of ChatGPT, almost the reverse happened, namely correction from the possible \emph{-r} Plural (though only rated 1.7 by German speakers) to the impossible plural.

\begin{LLM}
\textbf{EXP:} “Was ist der Plural von dem Wort Klot?”

\textbf{Chat GPT 3.5 (paid):} Der Plural von "Klot" ist "Klöter".

\end{LLM}

\begin{LLMTrans}
The plural of "Klot" is "Klöter".    
\end{LLMTrans}

\begin{LLM}
\textbf{EXP:} Bist du sicher?

\textbf{Chat GPT:} Entschuldigung, das war ein Missverständnis. Tatsächlich ist der Plural von "Klot" "Klötze".

\end{LLM}

\begin{LLMTrans}
Sorry, that was a misunderstanding. In fact, the plural of "Klot" is "Klötze".    
\end{LLMTrans}

In one case, it seemed that the LLM had knowledge of the nonsense word \emph{Pnähf} from the paper \citep{marcusetal95}.

\begin{LLM}
\textbf{EXP:} “Was ist der Plural von Pnähf?”

\textbf{Chat GPT:} Der Plural von "Pnähf" ist "Pnähf". Es handelt sich hierbei um einen Spaßbegriff oder ein Nonsenswort, für das es keinen fest definierten Plural gibt. Es wird oft verwendet, um auf die Eigenheiten der deutschen Sprache hinzuweisen, da Wörter im Deutschen normalerweise einen erkennbaren Plural haben.

\end{LLM}

\begin{LLMTrans}
The plural of "Pnähf" is "Pnähf". This is a joke term or a nonsense word for which there is no firmly defined plural. It is often used to point out the peculiarities of the German language, as words in German usually have a recognizable plural.    
\end{LLMTrans}

The full transcripts of our interactions with the seven LLMs are available as \cite{sauerlandetal25a}.

Two analyses show that as we predicted LLMs struggle to generalize German plural morphology correctly.
Figure~\ref{fig:llmcomparison} shows that all current LLMs produce forms that human speakers would rate in sum as implausible.  We separate here nonce nouns that rhyme with real nouns of Germans and non-rhyming nonce words.
The rhyming nonce words are often felt to be most natural with the same plural ending that one of the rhyming real nouns.
For comparison with human judgments, we give also the average rating reported for best rated form.
Most of the non-rhyming nouns (8 out of 12) are felt to be most natural with the \emph{-s} plural ending. With the exception of ChatGPT-4o, we find however that LLMs struggle more with the rhyming nouns.

\begin{figure*}
    \centering
    \includegraphics[width=0.9\linewidth]{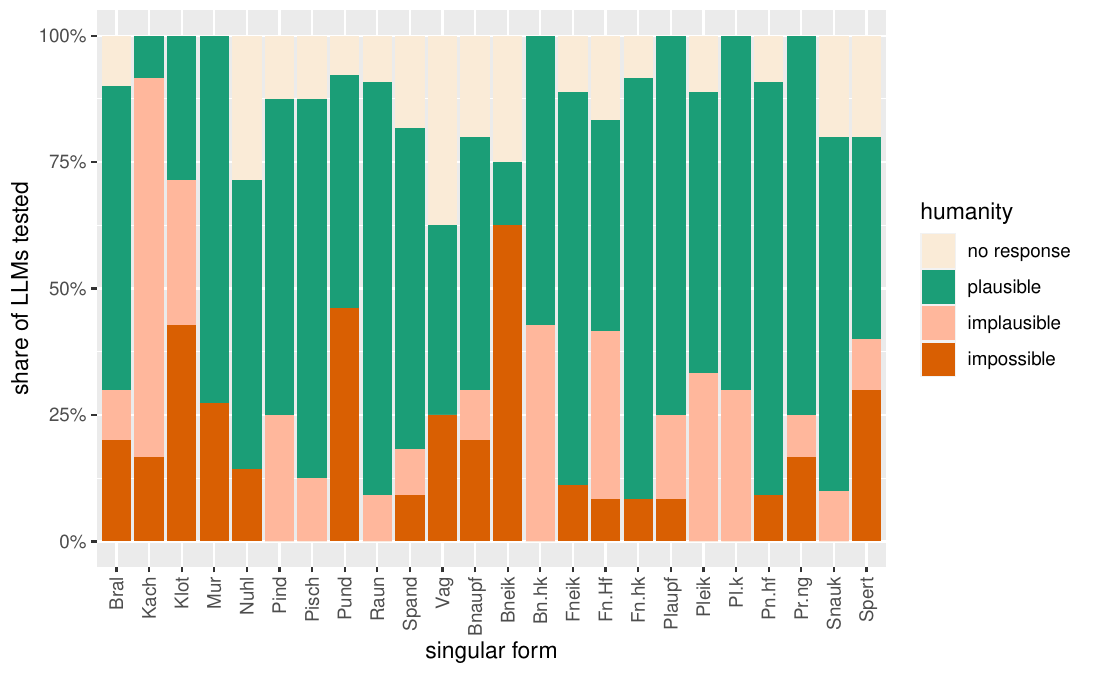}
    \caption{Item analysis of the nonce-noun plural task with rating with plausibility threshold 3.0. Stems on the left (\textit{Bral} to \textit{Bnaupf}) are rhyming while stems on the right are non-rhyming.}
    \label{fg:items}
\end{figure*}

We also investigated which of the 24 nonce nouns are particularly difficult for the seven LLMs.  Figure~\ref{fg:items} shows for each singular stem (12 rhyming stems on the left, 12 non-rhyming stems on the right) whether the plural forms generated by the LLMs would be considered plausible by German speakers, where we used the 3.0 as the threshold of plausibility.
The figure shows that  \textit{Kach} and \textit{Klot} are difficult for LLMs of the rhyming stems and Bneik of the non-rhyming stems.
With \textit{Kach} the preferred form by humans is the default plural \textit{Kach-s}, while most LLMs produce forms like \textit{Käche}, \textit{Kache}, and \textit{Kächer} assigning \textit{Kach} incorrectly to one of the limited classes.  For \textit{Klot}, several LLMs are driven to the frequent form \textit{Klötze}, which is morphologically impossible as the plural of \textit{Klot}. For \emph{Bneik}, the \emph{-s} plural (rating 4.3) and the \emph{-e} plural (rating 3.9) are reported to be plausible by \cite{marcusetal95}.

\section{Conclusion}

We conclude that the collaboration of computer scientists with linguists and cognitive scientists is likely to lead ideas for the improvement of LLMs. 
Our results resemble those of \cite{katzir23a} who also demonstrates that the linguistic abilities of LLMs exhibit predictable gaps.
As mentioned above, \cite{deepseek-aietal25} report that LLM-performance on reasoning tasks is substantially improved by applying  reinforcement learning with a set of mathematical and logical reasoning exercises.
This represents of promising start in our view and it might be fruitful to compile a more extensive set of exercises including linguistic ones for the training of LLMs -- we might call this a \textit{Curriculum for LLMs}.

At the same time, there is evidence that logical thinking is available to humans without explicit instruction.  One type of evidence comes from the fact that all languages including recently emerged languages provide the means to express logical quantification \cite{kocabetal22a}.  A second type of evidence comes from children's development. As mentioned above, some logical operations is found in prelinguistic infants and even negation seems present in the first year of life \citep{dautriche25}. 
Finally, \cite{sauerlandetal25} show that preschool children carry out higher order logical reasoning that requires a formal power exceeding first order logic.
In sum, it remains to be seen whether the differences between human language and LLM generated language can be overcome by enhanced LLM topologies and training methods.

\section*{Acknowledgements}

This project has received funding from the European Research Council (ERC) under the European Union’s Horizon 2020 research and innovation programme (grant agreement No 856421), project LeibnizDream, and from the German Research Foundation (DFG) under the ANR/DFG collaboration scheme, grant 510221551, project BooLL.
We thank Fabienne Salfner for her help with data management.

\section*{Data availability}

Full supporting data are openly available as \citep{sauerlandetal25a}.
The data content is fully documented in the file \texttt{README.md} contained in \citep{sauerlandetal25a}.

\bibliography{main}

\end{document}